\documentclass{article}
\usepackage{spconf,amsmath,graphicx}
\usepackage[utf8]{inputenc}

\usepackage{url}
\usepackage{booktabs}
\usepackage{amsfonts}
\usepackage{hyperref}
\usepackage{hyphenat}
\usepackage{nicefrac}
\usepackage{microtype}
\usepackage{enumitem}
\usepackage{paralist}
\usepackage{tabularx}

\newcommand{\wavetaco}{Wave-Tacotron}
\newcommand{\mcd}{MCD} %
\newcommand{\msd}{MSD} %

\newcommand{\mos}[2]{#1$\,\pm\,$#2}
\newcommand{\benchmark}[2]{#1}

\def\eqref#1{equation~\ref{#1}}

\def\1{\bm{1}}

\def\rvb{{\mathbf{b}}}
\def\rvc{{\mathbf{c}}}

\def\rve{{\mathbf{e}}}

\def\rvg{{\mathbf{g}}}

\def\rvx{{\mathbf{x}}}
\def\rvy{{\mathbf{y}}}
\def\rvz{{\mathbf{z}}}

\def\rmW{{\mathbf{W}}}

\DeclareMathAlphabet{\mathsfit}{\encodingdefault}{\sfdefault}{m}{sl}
\SetMathAlphabet{\mathsfit}{bold}{\encodingdefault}{\sfdefault}{bx}{n}

\newcommand{\krlmn}[5]{#2 & #4 & #5}
\newcolumntype{C}{r}
\newcolumntype{P}{r}

\title{
Wave-Tacotron:
Spectrogram-free
end-to-end %
text-to-speech
synthesis
}

\name{
Ron J. Weiss,
RJ Skerry-Ryan,
Eric Battenberg,
Soroosh Mariooryad,
Diederik P. Kingma
} %
\address{Google Research
    \\\texttt{\small \{ronw,rjryan,ebattenberg,soroosh,durk\}@google.com}}

\begin{document}
\ninept

\maketitle

\begin{abstract}
We describe a sequence-to-sequence neural network which directly generates speech waveforms from text inputs.
The architecture extends the Tacotron model by incorporating a normalizing flow into the autoregressive decoder loop.
Output waveforms are modeled as a sequence of non-overlapping fixed-length blocks, each one containing hundreds of samples.
The interdependencies of waveform samples within each block are modeled using the normalizing flow, enabling parallel training and synthesis.
Longer-term dependencies are handled autoregressively by conditioning each flow on preceding blocks.
This model can be optimized directly with maximum likelihood, without using intermediate, hand-designed features nor additional loss terms.
Contemporary state-of-the-art text-to-speech (TTS) systems use a cascade of separately learned models: one (such as Tacotron) which generates intermediate features (such as spectrograms) from text, followed by a vocoder (such as WaveRNN) which generates waveform samples from the intermediate features.
The proposed system, in contrast, does not use a fixed intermediate representation, and learns all parameters end-to-end.
Experiments show that the proposed model generates speech with quality approaching a state-of-the-art neural TTS system, %
 with significantly improved generation speed.
\end{abstract}

\section{Introduction}

Modern text-to-speech (TTS) synthesis systems, using deep neural networks trained on large quantities of data, are able to generate natural speech. %
TTS is a multimodal generation problem, since a text input can be realized in many ways, in terms of gross structure, e.g., with different prosody or pronunciation, and low level signal detail, e.g., with different, possibly perceptually similar, phase.
Recent systems
divide the task into
two steps, each tailored to one granularity.
First, a \emph{synthesis model} predicts intermediate audio features from text, typically spectrograms \cite{wang2017tacotron,shen2018natural}, vocoder features \cite{sotelo2017char2wav}, or linguistic features \cite{oord2016wavenet}, controlling the long-term  structure of the generated speech.
This is followed by a \emph{neural vocoder} which converts the features to time-domain waveform samples, filling in low-level signal detail.
These models are most commonly trained separately \cite{shen2018natural}, but can be fine-tuned jointly \cite{sotelo2017char2wav}.
Neural vocoder examples
include autoregressive models such as WaveNet \cite{oord2016wavenet} and WaveRNN \cite{kalchbrenner2018efficient},
fully parallel models using normalizing flows %
\cite{oord2018parallel,ping2019clarinet},
coupling-based flows which support efficient training and sampling %
\cite{kim2019flowavenet,prenger2019waveglow},
or hybrid flows which  model long-term structure in parallel but fine details autoregressively \cite{ping2020waveflow}.
Other approaches have eschewed probabilistic models using GANs \cite{kumar2019melgan,binkowski2020high}, or carefully constructed spectral losses %
\cite{arik2018fast,wang2019neural,gritsenko2020spectral}.

This two-step approach made %
end-to-end TTS tractable; however, it can complicate training and deployment, since attaining the highest quality generally requires fine-tuning \cite{sotelo2017char2wav} or training the vocoder on the output of the synthesis model \cite{shen2018natural} due to weaknesses in the latter.

End-to-end generation of waveform samples from text has been elusive, due to the difficulty
of efficiently modeling strong temporal dependencies in the waveform, e.g., phase, which must be consistent over time.
Sample-level autoregressive vocoders handle such dependencies by conditioning generation of each waveform sample on all previously generated samples.
Due to their highly sequential nature, they are inefficient to sample from on modern parallel hardware.

In this paper we integrate a flow-based vocoder into a Tacotron-like block-level autoregressive model of speech waveforms, conditioned on a sequence of characters or phonemes. %
The sequence-to-sequence decoder attends to the input text, and produces conditioning features for a normalizing flow which generates waveform blocks.

The output waveform is modeled as a sequence of fixed-length blocks, each containing hundreds of samples.
Dependencies between samples within each block are modeled using a normalizing flow, enabling parallel training and synthesis.
Long-term dependencies are modeled autoregressively by conditioning on previous blocks.
The resulting model solves a more difficult task than text-to-spectrogram generation since it must synthesize the fine-time structure in the waveform (ensuring that edges of successive output blocks line up correctly) as well as produce coherent long-term structure (e.g., prosody and semantics).
Nevertheless, we find that
the model is able to align the text and generate high fidelity speech without framing artifacts.

Recent work has integrated normalizing flows with sequence\hyp{}to\hyp{}sequence TTS models, to enable a non\hyp{}autoregressive decoder %
\cite{miao2020flow}, or improve modelling of the intermediate mel spectrogram %
\cite{valle2020flowtron,kim2020glow}.
\cite{ren2020fastspeech} and \cite{donahue2020end} %
also propose end-to-end TTS models which directly generate waveforms, but rely on spectral losses and mel spectrograms for alignment.
In contrast, we avoid spectrogram generation altogether and use a normalizing flow to directly model time-domain waveforms, in a fully end-to-end approach, simply maximizing likelihood.

\section{Wave-Tacotron Model}
\label{sec:model}
\begin{figure*}[t]
  \centering
   \includegraphics[width=0.99\linewidth]{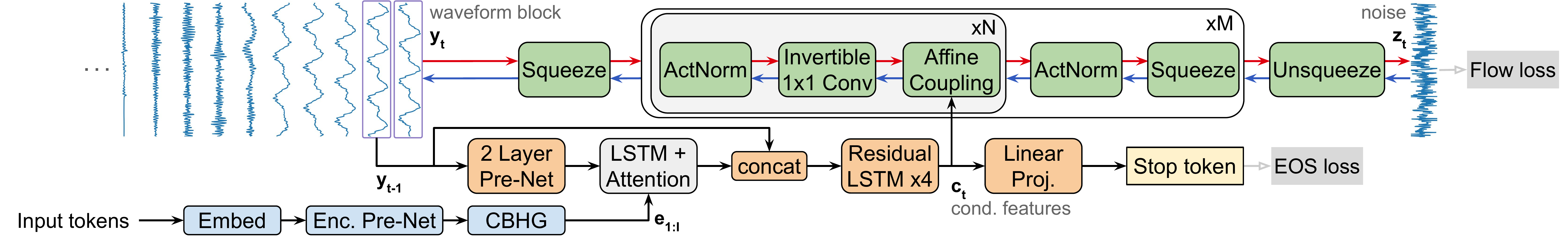}
   \vskip-1.5ex
   \caption{\wavetaco{} architecture.
The text encoder (blue, bottom) output is passed to the decoder (orange, center) to  produce conditioning features for a normalizing flow (green, top) which models non-overlapping waveform segments spanning $K$ samples for each decoder step.
 During training (red arrows) the flow converts a block of waveform samples to noise; during generation (blue arrows) its inverse is used to generate waveforms from a random noise sample.  The model is trained by maximizing the likelihood using the flow and EOS losses (gray).}
   \label{fig:architecture}
\end{figure*}

\wavetaco{} extends the Tacotron attention-based sequence-to-sequence model, generating blocks of non-overlapping waveform samples instead of spectrogram frames.
See Figure~\ref{fig:architecture} for an overview. %
The encoder is a CBHG~\cite{wang2017tacotron}, which encodes a sequence of $I$ token (character or phoneme) embeddings $\rvx_{1:I}$. %
The text encoding is passed to a block-autoregressive decoder %
using attention,
producing conditioning features $\rvc_t$ for each output step $t$.
A normalizing flow \cite{rezende2015variational} uses these features to sample an output waveform for that step, $\rvy_t$:
\begin{align}
  \rve_{1:I} &= \textbf{encode}(\rvx_{1:I}) \\
  \rvc_t &= \textbf{decode}(\rve_{1:I},\, \rvy_{1:t-1}) \\
  \rvy_t &= \rvg(\rvz_t; \rvc_t) \quad \text{where} \quad \rvz_t \sim \mathcal{N}(\mathbf{0}, \mathbf{I})
  \label{eq:flow_sampling} %
\end{align}
The flow $\rvg$ 
converts a random vector $\rvz_t$ sampled from a Gaussian into a waveform block.
A linear classifier on $\rvc_t$ computes the probability $\hat{s}_t$ that $t$ is the final output step (called a \emph{stop token} following \cite{shen2018natural}):
\begin{align}
  \hat{s}_t &= \sigma(\rmW_s \rvc_t + \rvb_s)
\end{align}
where $\sigma$ is a sigmoid nonlinearity.
Each output $\rvy_t \in \mathbb{R}^K$ by default comprises 40 ms of speech: $K=960$ at a 24 kHz sample rate.
The setting of $K$ controls the trade-off between parallelizability through normalizing flows, and sample quality through autoregression.

The network structure follows \cite{wang2017tacotron,shen2018natural}, with minor modifications to the decoder.
We use location-sensitive attention~\cite{chorowski2015attention}, %
which was more stable than the non-content-based GMM attention from \cite{battenberg2020location}.
We replace ReLU activations with tanh in the pre-net.
Since we sample from the normalizing flow in the decoder loop, %
we do not need to apply pre-net dropout when sampling as in \cite{shen2018natural}.
We also do not use a post-net \cite{shen2018natural}. 
Waveform blocks generated at each decoder step are simply concatenated %
to form the final signal.
Finally, we add a skip connection over the decoder pre-net and attention layers, to give the flow direct access to the samples directly preceding the current block. This is essential to avoid discontinuities at block boundaries.

Similar to ~\cite{wang2017tacotron}, the size $K$ of $\rvy_t$ is controlled by a \emph{reduction factor} hyperparameter $R$, where $K = 320 \cdot R$,
and $R = 3$ by default. %
Like~\cite{wang2017tacotron}, the autoregressive input to the decoder consists of only the final $K/R$ samples of the output from the previous step $\rvy_{t-1}$.%

\subsection{Flow}

The normalizing flow $\rvg(\rvz_t; \rvc_t)$ is a composition of invertible transformations which maps a noise sample drawn from a spherical Gaussian to a waveform segment.
Its representational power comes from composing many simple functions.
During training, the inverse $\rvg^{-1}$ maps the target waveform to a point under the spherical Gaussian whose density is easy to compute.

\begin{figure}[t]
  \centering
   \includegraphics[width=0.99\linewidth]{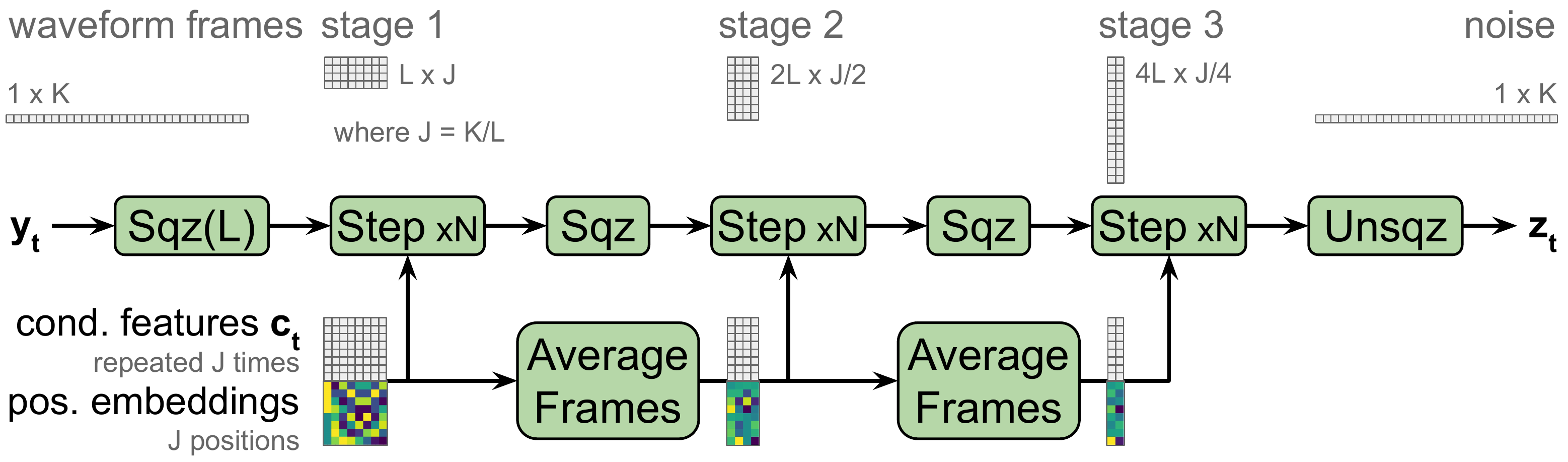}
   \vskip-1.5ex
   \caption{Expanded flow with 3 stages.
   top: Activation shapes within each stage.
   bottom: Conditioning features are stacked with position embeddings for each of the $J$ timesteps in the first stage.
   Adjacent frame pairs are averaged together when advancing between stages, harmonizing with  squeezes but without increasing dimension.}
   \label{fig:flow_stages}
\end{figure}

Since training and generation require efficient density computation and sampling, respectively, $\rvg$ is constructed using affine coupling layers \cite{dinh2015nice}.
Our normalizing flow is a one-dimensional variant of Glow~\cite{kingma2018glow}, similar to \cite{kim2019flowavenet}.
During training, the input to $\rvg^{-1}$, $\rvy_t \in \mathbb{R}^{K}$, is first squeezed into a sequence of $J$ frames, each with dimension $L = 10$.
The flow is divided into $M$ \emph{stages}, each operating at a different temporal resolution, following the multiscale configuration described in \cite{dinh2016density}.
This is implemented by interleaving \emph{squeeze} operations which reshape the sequence between each stage, halving the number of timesteps and doubling the dimension, as illustrated in Figure~\ref{fig:flow_stages}.
After the final stage, an unsqueeze operation flattens the output $\rvz_t$ to a vector in $\mathbb{R}^K$.
During generation, to transform $\rvz_t$ into $\rvy_t$ via $\rvg$, each operation is inverted and their order is reversed.

Each stage is further composed of $N$ flow \emph{steps} as shown in the top of Figure~\ref{fig:architecture}.
We do not use data-dependent initialization for our ActNorm layers as in \cite{kingma2018glow}.
The affine coupling layers emit transformation parameters using a 3-layer 1D convnet with kernel sizes 3, 1, 3, with 256 channels.
Through the use of squeeze layers, coupling layers in different stages use the same kernel widths yet have different effective receptive fields relative to $\rvy_t$.
The default configuration uses $M=5$ %
and $N=12$, for a total of 60 steps.

The conditioning signal $\rvc_t$ consists of a single vector for each decoder step, which must %
encode the structure of $\rvy_t$ across hundreds of samples.
Since the flow internally treats $\rvz_t$ and $\rvy_t$ as a sequence of $J$ frames, we find it helpful to upsample $\rvc_t$ to match this framing.
Specifically, we replicate $\rvc_t$ $L$ times and concatenate 
sinusoidal position embeddings to each time step, similar to \cite{vaswani2017attention}, albeit with linear spacing between frequencies.
The upsampled conditioning features are appended to the inputs at each coupling layer.

To avoid issues associated with fitting a real-valued distribution to discrete waveform samples, we adapt the approach in \cite{uria2013rnade} by quantizing samples into $[0, 2^{16}-1]$ levels, dequantizing by adding uniform noise in $[0, 1]$, and rescaling back to $[-1, 1]$.
We also found it helpful to pre-emphasize the waveform, whitening the signal spectrum using an FIR highpass filter with a zero at 0.9. 
The generated signal is de-emphasized at the output of the model.
This pre-emphasis can be thought of as a flow without any learnable parameters, acting as an inductive bias to place higher weight on high frequencies.
Although not critical, we found that this improved subjective listening test results.

As is common practice with generative flows, we find it helpful to reduce the temperature of the prior distribution when sampling.
This amounts to simply scaling the prior covariance of $\rvz_t$ in equation \ref{eq:flow_sampling} by a factor $T^2 < 1$.
We find $T=0.7$ to be a good default setting.

During training, the inverse of the flow, $\rvg^{-1}$, maps the target waveform $\rvy_t$ to the corresponding $\rvz_t$.
Conditioning features $\rvc_t$ are calculated using teacher-forcing of the sequence-to-sequence %
network.
The negative log-likelihood objective of the flow at step $t$ is \cite{dinh2015nice}:
\begin{align}
  \mathcal{L}_{\text{flow}}(\rvy_t) %
      &= -\log p(\rvy_t | \rvx_{1:I}, \rvy_{1:t-1})
       = -\log p(\rvy_t | \rvc_t) \nonumber \\
      &= -\log \mathcal{N}(\rvz_t; \mathbf{0}, \mathbf{I})
       -\log \left|\det(\partial\rvz_t / \partial\rvy_t)\right| \label{eq:loss_flow} %
\end{align}
where $\rvz_t = \rvg^{-1}(\rvy_t; \rvc_t)$.
A binary cross entropy loss is also used on the stop token classifier:
\begin{align}
  \mathcal{L}_{\text{eos}}(s_t)
    &= -\log p(s_t | \rvx_{1:I}, \rvy_{1:t-1}) = -\log p(s_t | \rvc_t) \nonumber \\
    &= -s_t\log \hat{s}_t - (1-s_t) \log(1-\hat{s}_t) \label{eq:loss_eos}
\end{align}
where $s_t$ is the ground truth stop token label, indicating whether $t$ is the final step in the utterance.
In practice we zero-pad the signal with several blocks labeled with $s_t=1$ in order to provide a better balanced training signal to the classifier.
We use the mean of eq.~\ref{eq:loss_flow} and \ref{eq:loss_eos} across all decoder steps $t$ as the overall loss. 

\vspace{-0.1ex}
\subsection{Flow vocoder}
\label{sec:flowcoder}

As a baseline, we experiment with a similar flow network in a fully feed-forward vocoder context, which generates waveforms from mel spectrograms as in \cite{kim2019flowavenet,prenger2019waveglow}.
We follow the architecture in Fig.~\ref{fig:flow_stages}, with flow frame length $L=15$, $M=6$ stages and $N=10$ steps per stage.
Input mel spectrogram features are encoded using a 5-layer dilated convnet conditioning stack using 512 channels per layer, similar to %
that in \cite{kalchbrenner2018efficient}.
Features are upsampled to match the flow frame rate of 1600 Hz, and concatenated with sinusoidal embeddings to indicate position within each feature frame, as above.
This model is highly parallelizable, since it does not make use of autoregression.

\section{Experiments}

We experiment with two single-speaker datasets:
A proprietary dataset containing about 39 hours of speech, sampled at 24~kHz, from a professional female voice talent which was used in previous studies \cite{oord2016wavenet,wang2017tacotron,shen2018natural}.
In addition, we experiment with the public LJ speech dataset \cite{ito2017ljspeech} of audiobook recordings,
using the same splits as \cite{battenberg2020location}: 22 hours for training and a held-out 130 utterance subset for evaluation.
We upsample the 22.05~kHz audio to 24~kHz for consistency with the first dataset.
Following common practice, we use a text normalization pipeline and pronunciation lexicon to map input text into a sequence of phonemes.
For a complete end-to-end approach, we also train models that take characters, after text normalization, as input.

For comparison we train three baseline end-to-end TTS systems:
The first is a Tacotron model predicting mel spectrograms with $R=2$, corresponding to 25ms per decoder step.
We pair it with a separately trained WaveRNN \cite{kalchbrenner2018efficient} vocoder to generate waveforms, similar to Tacotron~2 \cite{shen2018natural}.
The WaveRNN uses a 768-unit GRU cell, a 3-component mixture of logistics output layer, and a conditioning stack similar to the flow vocoder described in Section~\ref{sec:flowcoder}.
In addition, we train a flow vocoder (Flowcoder) similar to \cite{prenger2019waveglow,kim2019flowavenet} to compare with WaveRNN.
Finally, as a lower bound on performance, we jointly train a Tacotron model with a post-net (Tacotron-PN),
consisting of a 20-layer non-causal WaveNet stack split into two dilation cycles. This converts mel spectrograms output by the decoder to full linear frequency spectrograms which are inverted to waveform samples using 100 iterations of the Griffin-Lim algorithm \cite{griffin1984signal},
 similar to %
\cite{wang2017tacotron}.

\wavetaco{} models are trained using the Adam optimizer for 500k steps on 32 Google TPUv3 cores, with batch size 256 for the proprietary dataset and 128 for LJ, whose average speech duration is longer.

Performance is measured using subjective listening tests, crowd-sourced via a pool of native speakers listening with headphones.
Results are measuring using the mean opinion score (MOS) rating naturalness of generated speech on a ten-point scale from 1 to 5.
We also compute several objective metrics:
\begin{inparaenum}[(1)]
\item
Mel cepstral distortion (\mcd{}), the root mean squared error against the ground truth signal, computed on 13 dimensional MFCCs \cite{kubichek1993mel} using dynamic time warping \cite{berndt1994using} to align features computed on the synthesized audio to those computed from the ground truth.
MFCC features are computed from an 80-channel log-mel spectrogram using a 50ms Hann window and hop of 12.5ms.
\item
Mel spectral distortion (\msd{}), which is the same as \mcd{} but applied to the log-mel spectrogram magnitude instead of cepstral coefficients.
This captures harmonic content which is explicitly removed in \mcd{} due to liftering.
\item
Character error rate (CER) computed after transcribing the generated speech with the Google speech API, as in \cite{battenberg2020location}.
This is a rough measure of intelligibility, invariant to some types of acoustic distortion.
In particular CER is sensitive to robustness errors caused by stop token or attention failures.
\end{inparaenum}
Lower values are better for all metrics except MOS.
As can be seen below, although \msd{} and \mcd{} tend to negatively correlate with subjective MOS, we primarily focus on MOS.

\begin{table}[t]
\centering
\setlength{\tabcolsep}{1.4pt}
\begin{tabular}{@{}l@{\hspace{2pt}}ll@{\hspace{-1pt}}rrC@{\hspace{6pt}}r@{}}
\toprule
Model & Vocoder & Input & \mcd{} & \msd{} & CER & MOS \\
\midrule
Ground truth &\hspace{1pt}--& -- & -- & -- & 8.7 & \mos{4.56}{0.04} \\
Tacotron-PN & Griffin-Lim & char & 5.87 & 11.40 & 9.0 & \mos{3.68}{0.08}\\
Tacotron-PN & Griffin-Lim & phoneme & 5.83 & 11.14 & 8.6 & \mos{3.74}{0.07} \\
Tacotron & WaveRNN & char & 4.32 & 8.91 & 9.5 & \mos{4.36}{0.05} \\
Tacotron & WaveRNN & phoneme & 4.31 & 8.86 & 9.0 & \textbf{\mos{4.39}{0.05}}\\
Tacotron & Flowcoder & char & 4.22  & 8.81 & 11.3 & \mos{3.34}{0.07} \\
Tacotron & Flowcoder & phoneme & \textbf{4.15} & \textbf{8.69} & 10.6 & \mos{3.31}{0.07} \\
\wavetaco{} &\hspace{1pt}--& char & 4.84 & 9.47  & 9.4 & \mos{4.07}{0.06} \\
\wavetaco{} &\hspace{1pt}--& phoneme & 4.64 & 9.24 & 9.2 & \mos{4.23}{0.06} \\
\bottomrule
\end{tabular}
\caption{TTS performance on the proprietary single speaker dataset.}
\label{tbl:tts_hol}
\end{table}

Results on the proprietary dataset are shown in Table~\ref{tbl:tts_hol}, comparing performance using phoneme or character inputs.
Unsurprisingly, all models perform slightly better on phoneme inputs.
Models trained on character inputs are more prone to pronunciation errors, since they must learn the mapping directly from data, whereas phoneme inputs already capture the intended pronunciation.
We note that the performance gap between input types is typically not very large for any given system, indicating the feasibility of end-to-end training.
However, the gap is largest for direct waveform generation with \wavetaco{}, indicating perhaps that the model capacity is being used to model detailed waveform structure instead of pronunciation.

Comparing MOS across  different systems, in all cases \wavetaco{} performs substantially better than the Griffin-Lim baseline, but not as well as Tacotron + WaveRNN.
Using the flow as a vocoder results in \mcd{} and \msd{} on par with Tacotron + WaveRNN, but increased CER %
and
low MOS, where raters noted robotic and gravelly sound quality.
The generated audio contains artifacts on short timescales, which have little impact on spectral metrics computed with 50ms windows.
In contrast, \wavetaco{} has higher \mcd{} and \msd{} than the Tacotron + WaveRNN baseline, but similar CER and only slightly lower MOS.
The significantly improved fidelity of \wavetaco{} over Flowcoder indicates the utility of autoregressive conditioning, which resolves some uncertainty about the fine-time structure within the signal
that Flowcoder must implicitly learn to model.
However, fidelity still falls short of the WaveRNN baseline.

Table~\ref{tbl:tts_ljspeech} shows a similar ranking on the LJ dataset.
However, the MOS gap between \wavetaco{} and the baseline is larger.
\wavetaco{} and Flowcoder \msd{} are lower, suggesting that the flows are more prone to fine-time scale artifacts.
The LJ dataset is much smaller, leading to the conclusion that fully end-to-end training likely requires more data than the two-step approach.
Note also that hyperparameters were tuned on the proprietary dataset.
Sound examples are available
at \url{https://google.github.io/tacotron/publications/wave-tacotron}.\footnote{
We include samples from an unconditional \wavetaco{}, removing the encoder and attention, which is capable of generating 
coherent  syllables. %
}

\begin{table}[t]
\centering
\setlength{\tabcolsep}{1.5pt}
\begin{tabular}{@{}lllrrC@{\hspace{7pt}}r@{}}
\toprule
Model & Vocoder & Input & \mcd{} &  \msd{} & CER & MOS \\
\midrule
Ground truth & -- & -- & -- & -- & 7.8 & 4.51$\,\pm\,$0.05 \\
Tacotron-PN & Griffin-Lim & char & 7.20 & 12.39 & 7.4 & \mos{3.26}{0.11} \\
Tacotron & WaveRNN & char & 6.69 & 12.23 & 6.1 & \textbf{\mos{4.47}{0.06}} \\
Tacotron & Flowcoder & char & \textbf{5.80} & 11.61 & 13.0 & \mos{3.11}{0.10}\\
\wavetaco{} & -- & char & 6.87 & \textbf{11.44} & 9.2 & \mos{3.56}{0.09} \\
\bottomrule
\end{tabular}
\caption{TTS performance on LJ Speech with character inputs.}
\label{tbl:tts_ljspeech}
\end{table}

\subsection{Generation speed} \label{sec:speed}

\begin{table}[t]
\centering
\begin{tabular}{@{}l@{\hspace{4pt}}r@{\hspace{6pt}}lrP@{}}
\toprule
Model & $R$ & Vocoder & TPU & CPU  \\
\midrule
Tacotron-PN &2 & Griffin-Lim, 100 iterations & \benchmark{0.14}{0.002} & \benchmark{0.88}{0.008} \\
Tacotron-PN &2 & Griffin-Lim, 1000 iterations & \benchmark{1.11}{0.001}  & \benchmark{7.71}{0.055} \\
Tacotron    &2 & WaveRNN & \benchmark{5.34}{0.004} & \benchmark{63.38}{1.619}  \\
Tacotron    &2 & Flowcoder & \benchmark{0.49}{0.004} & \benchmark{0.97}{0.007} \\
\wavetaco{} &1 & -- & \benchmark{0.80}{0.003} & \benchmark{5.26}{1.246} \\
\wavetaco{} &2 & -- & \benchmark{0.64}{0.003} & \benchmark{3.25}{0.012} \\
\wavetaco{} &3 & -- & \benchmark{0.58}{0.002} & \benchmark{2.52}{0.009} \\
\wavetaco{} &4 & -- & \benchmark{0.55}{0.003} & \benchmark{2.26}{0.008} \\
\bottomrule
\end{tabular}
\caption{
Generation speed in seconds, comparing one TPU v3 core to a 6-core Intel Xeon W-2135 CPU, 
generating 5 seconds of speech conditioned on 90 input tokens, batch size 1. Average of 500 trials.}
\label{tbl:generation_speed}
\end{table}

Table~\ref{tbl:generation_speed} compares end-to-end generation speed of audio from input
text on TPU and CPU.
The baseline Tacotron+WaveRNN has the slowest generation speed, slightly slower than real-time on TPU.
This is a consequence of the autoregressive sample-by-sample generation process for WaveRNN (note that we do not use a custom kernel to speed up sampling as in \cite{kalchbrenner2018efficient}).
Tacotron-PN is substantially faster, as is Tacotron + Flowcoder, which uses a massively parallel vocoder network, processing the full spectrogram generated by Tacotron in parallel.
\wavetaco{} with the default $R=3$ is about 17\% slower than Tacotron + Flowcoder on TPU, and about twice as fast as Tacotron-PN using 1000 Griffin-Lim iterations.
On its own, spectrogram generation using Tacotron is quite efficient, as demonstrated by the fast speed of Tacotron-PN with 100 Griffin-Lim iterations.
On parallel hardware, and with sufficiently large decoder steps, the additional overhead in the decoder loop of the \wavetaco{} flow, essentially a very deep convnet, is small over the fully parallel Flowcoder.

\subsection{Ablations} \label{sec:ablations}

\begin{table}[t]
\centering
\setlength{\tabcolsep}{3.2pt} %
\begin{tabular}{@{}l@{\hspace{4pt}}
rrr
rrC%
r@{}}
\toprule
Model & \krlmn{$K$}{$R$}{$L$}{$M$}{$N$} &\mcd{} &  \msd{} & CER & MOS \\
\midrule
Base $T=0.8$ & \krlmn{320}{3}{10}{5}{12} & 4.43 & 9.20 & 9.0 & \mos{4.01}{0.06}\\
\addlinespace
$T=0.6$ & \krlmn{320}{3}{10}{5}{12} & 5.02 & 9.81 & 9.1 & 	\mos{4.12}{0.06}\\
$T=0.7$ & \krlmn{320}{3}{10}{5}{12} & 4.71 & 9.44 & 9.2 & \mos{4.16}{0.06}\\
$T=0.9$ & \krlmn{320}{3}{10}{5}{12} & 4.34 & 9.40 & 9.1 & \mos{3.77}{0.07}\\
\addlinespace
no pre-emphasis & \krlmn{320}{3}{10}{5}{12} & 4.46 & 9.27 & 9.2 & \mos{3.85}{0.06} \\
no position emb. & \krlmn{320}{3}{10}{5}{12} & 4.64 & 9.48 & 9.3 & \mos{3.70}{0.07} \\
no skip connection %
    & \krlmn{320}{3}{10}{5}{12} & 5.67 & 10.54 & 13.2 & --\\
\addlinespace
128 flow channels & \krlmn{320}{3}{10}{5}{12} & 4.40 & 9.29 & 9.4 & \mos{3.31}{0.07} \\
30 steps, 5 stages & \krlmn{320}{3}{10}{5}{$\;\,$6} & 4.33 & 9.17 & 9.3 & \mos{3.11}{0.07} \\
60 steps, 4 stages & \krlmn{320}{3}{10}{4}{15} & 4.40 & 9.35  & 8.8 & \mos{3.50}{0.07} \\
60 steps, 3 stages & \krlmn{320}{3}{10}{3}{20} & 4.51 &  9.45 & 9.8 & \mos{2.44}{0.07} \\
\addlinespace
$K=320$ (13.33 ms) & \krlmn{320}{1}{10}{5}{12} & 5.35 & 9.83 & 11.3 & \mos{4.05}{0.06} \\
$K=640$ (26.67 ms) & \krlmn{320}{2}{10}{5}{12} & 4.51 & 9.15 & 8.9 & \mos{4.06}{0.06} \\
$K=1280$ (53.3 ms) & \krlmn{320}{4}{10}{5}{12} & 4.42 & 9.43 & 9.3 & \mos{3.55}{0.07} \\
\bottomrule
\end{tabular}
\caption{Ablations on the proprietary dataset using phoneme inputs and a shallow decoder residual LSTM stack of 2 layers with 256 units.
Unless otherwise specified, samples are generated using $T=0.8$.}
\label{tbl:tts_hol_ablations}
\end{table}

Finally, Table~\ref{tbl:tts_hol_ablations} compares several \wavetaco{} variations, using a shallower 2-layer decoder LSTM. %
As shown in the top, the optimal sampling temperature is $T=0.7$.
Removing pre-emphasis, position embeddings in the conditioning stack, or the decoder skip connection, all hurt performance.
Removing the skip connection is particularly deleterious, leading to clicking artifacts caused by discontinuities at block boundaries which significantly increase \mcd{}, \msd{}, and CER, to the point where we chose not to run listening tests.

Decreasing the size of the flow network by halving the number of coupling layer channels to 128, or decreasing the number of stages $M$ or steps per stage $N$, all significantly reduce quality in terms of MOS.
Finally, we find that reducing $R$ does not hurt MOS, although it slows down generation (see Table~\ref{tbl:generation_speed}) by increasing the number of autoregressive decoder steps needed to generate the same audio duration.
Increasing $R$ to 4 significantly reduces MOS due to poor prosody, with raters commenting on the unnaturally fast speaking rate.
This suggests a trade-off between parallel generation (by increasing $K$, decreasing the number of decoder steps) and naturalness. 
Feeding back more context in the autoregressive decoder input or increasing the capacity of the decoder LSTM stack would likely mitigate these issues; we leave such exploration %
for future work.

\section{Discussion}

We have proposed a model for end-to-end (normalized) text-to-speech waveform synthesis, incorporating a normalizing flow into the autoregressive Tacotron decoder loop.
\wavetaco{} directly generates high quality speech waveforms, conditioned on text,  using a single model
without a separate vocoder.
Training does not require complicated losses on hand-designed spectrogram or other mid-level features, but simply maximizes the likelihood of the training data.
The hybrid model structure combines the simplicity of attention-based TTS models with the parallel generation capabilities of a normalizing flow to generate waveform samples directly.

Although the network structure exposes the output block size $K$ as a hyperparameter, the decoder remains fundamentally autoregressive, requiring sequential generation of output block.
This puts the approach at a disadvantage compared to recent advances in parallel TTS \cite{ren2019fastspeech,yu2019durian},
unless the output step size can be made very large.
Exploring the feasibility of adapting the proposed model to fully parallel TTS generation remains an interesting direction for future work.
It is possible that the separation of concerns inherent in the two-step %
factorization of the TTS task, i.e., separating responsibility for high fidelity waveform generation from text-alignment and longer term prosody modeling, leads to more efficient models in practice.
Separate training of the two models also allows for the possibility of training them on different data, e.g., training a vocoder on a larger corpus of untranscribed speech.
Finally, it would be interesting to explore more efficient alternatives to flows in a similar text-to-waveform setting, e.g., diffusion probabilistic models \cite{chen2020wavegrad}, or GANs \cite{binkowski2020high,donahue2020end}, which can be more easily optimized in a mode-seeking fashion that is likely to be more efficient than modeling the full data distribution.

\section{Acknowledgements}
The authors thank
Jenelle Feather for initial work integrating a flow into Tacotron,
Rif A. Saurous for facilitating the genesis of this work, and
Tom Bagby, David Kao, Matt Shannon, Daisy Stanton, Yonghui Wu, Heiga Zen, and Yu Zhang for helpful discussions and feedback.

\bibliographystyle{IEEEbib}
\bibliography{refs}

\appendix
\section{Autoregressive input}

As described in Section~\ref{sec:model}, \wavetaco{} uses the reduction factor mechanism from \cite{wang2017tacotron} which truncates the autoregressive input to the decoder at each step. %
Although this works well in practice, we note that it has the unusual effect of breaking the model's autoregressive decomposition  since segments at the beginning of each output block are never seen by the decoder in subsequent steps.
In order to validate that this does not hurt generation quality, we trained a variant of the model on the proprietary dataset (with phoneme inputs) 
using the same frame size but no reduction factor truncation, i.e., using block size $K=960$ with $R=1$.
We conducted a subjective listening test measuring  preference between samples generated with the default \wavetaco{} model using $R=3$ from Table~\ref{tbl:tts_hol}, and the variant without a reduction factor.
Raters were presented with the two samples side-by-side and asked to rate their preference between them on a 7-point scale, scaled between -3 and 3 \cite{shen2018natural}.
The average score of \mos{0.028}{0.126} revealed no preference between  models, indicating that reduction factor mechanism does not impair generation quality.

\section{Sample variability}

\begin{figure}[t!]
  \centering
   \includegraphics[width=0.85\linewidth]{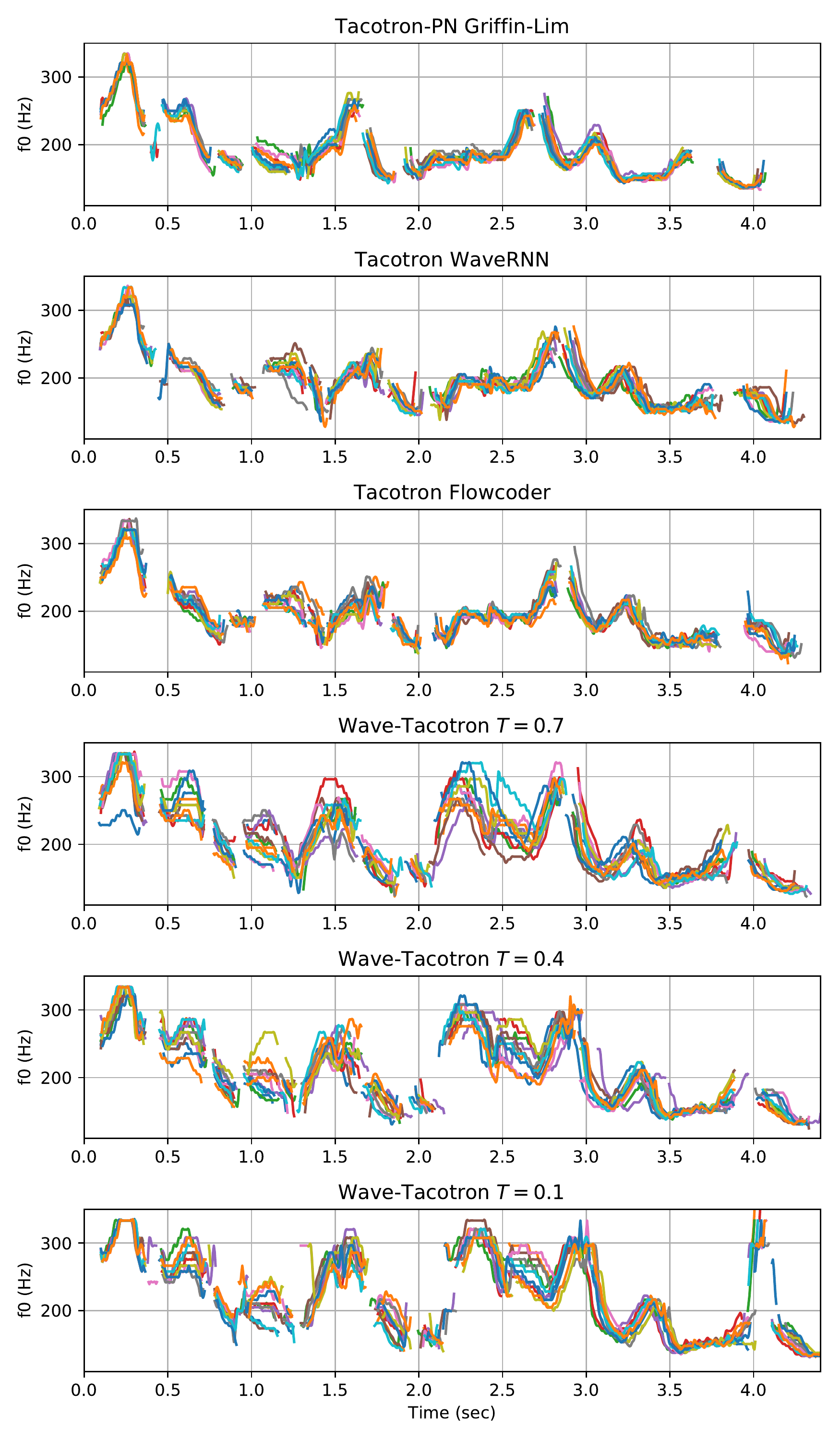}
   \vskip-2.5ex
   \caption{Visualizing sample variability across TTS systems.
   Each subplot shows pitch tracks computed from 12 independent samples generated from the same input text.
   top rows: Different TTS systems, bottom rows: \wavetaco{} varying the sampling temperature $T$.
   }
   \label{fig:pitch_tracks}
\end{figure}

Figure~\ref{fig:pitch_tracks} visualizes pitch tracks corresponding to several independent samples generated from the same input text, across the TTS systems from Table~\ref{tbl:tts_hol} taking phone inputs.
Overlaying pitch tracks for several samples demonstrates the degree of variation in terms of timing and pitch trajectories generated by a given system.

As shown in top three rows of Figure~\ref{fig:pitch_tracks}, spectrogram-based Tacotron models generate speech with very consistent pitch tracks across samples.
In contrast, \wavetaco{} has significantly more variation, e.g., evidenced by the differing trajectories in the two vowels in the first 0.7 seconds and throughout the utterance,
following similar Flow-based TTS models \cite{valle2020flowtron}.
Similar trends remain apparent at relatively low sampling temperatures, although the degree of variation decreases with temperature (bottom rows).
This suggests a potential application of \wavetaco{} as a data augmentation method for speech recognition \cite{sun2020prosody_prior}.

Since it is trained to maximize likelihood, \wavetaco{} captures the richer multimodal distribution of the speech in the training data.
In contrast, the baseline Tacotron construction, which minimizes a  regression loss and has limited non-determinism during generation (due to dropout during eval \cite{shen2018natural}), naturally tends to regress to the mean and therefore collapses to a single "prosodic mode".
Several recent papers have explored more complex extensions which are able to better capture richer distributions of prosody, based on prosody embeddings extracted from a reference utterance \cite{skerry2018towards}, or incorporating VAE conditioning into the decoder \cite{hsu2019hierarchical}.
In contrast to these approaches, \wavetaco{} does not depend on more complicated modeling assumptions.
The generated prosody is controlled by the noise samples passed into the flow $\rvz_t$.
The downside is that these noise values are more difficult to interpret and control compared to prosody embedding-based approaches, due to their extremely high dimensionality and very localized effects due to the one-to-one correspondence between dimensions of $\rvz_t$ and the output waveform $\rvy_t$.
Since the model is trained to generate waveforms directly, $\rvz_t$ also entangles mid-level prosodic features with low level signal detail such as phase and precise timing, for which precise control or interpretability are unnecessary.
We leave exploration of techniques for interpretable control as an area for future work.

\end{document}